\documentclass[extendedabs]{recpad2k}

\title{Chain-of-Anomaly Thoughts with Large Vision-Language Models}

\addauthor{Pedro Domingos}{pa.domingos@campus.fct.unl.pt}{1}
\addauthor{João Pereira}{jaca.pereira@campus.fct.unl.pt}{1}
\addauthor{Vasco Lopes}{vasco.lopes@ubi.pt}{3}
\addauthor{João Neves}{jcneves@ubi.pt}{2}
\addauthor{David Semedo}{df.semedo@fct.unl.pt}{1}

\addinstitution{
 NOVALINCS, NOVA School of Science and Technology\\
 2829-516 Caparica, PT\\
}
\addinstitution{
 NOVALINCS, Universidade da Beira Interior\\
 Rua Marquês D’Ávila e Bolama\\
 6200-001, Covilhã, PT
}
\addinstitution{
 DeepNeuronic
 Covilhã, PT
}

\runninghead{Domingos, Pereira, Lopes, Semedo, Neves}{RECPAD Author Guidelines}

\usepackage[normalem]{ulem}

\begin{document}

\maketitle

\begin{abstract}
Automated video surveillance with Large Vision-Language Models is limited by their inherent bias towards normality, often failing to detect crimes. While Chain-of-Thought reasoning strategies show significant potential for improving performance in language tasks, the lack of inductive anomaly biases in their reasoning further steers the models towards normal interpretations. To address this, we propose Chain-of-Anomaly-Thoughts (CoAT), a multi-agent reasoning framework that introduces inductive criminal bias in the reasoning process through a final, anomaly-focused classification layer. Our method significantly improves Anomaly Detection, boosting F1-score by 11.8 p.p. on challenging low-resolution footage and Anomaly Classification by 3.78 p.p. in high-resolution videos. 
\end{abstract}


\section{Introduction}
\label{sec:intro}

\renewcommand{\thefootnote}{}
\footnotetext{This work is supported by NOVA LINCS ref. UIDB/04516/2020 (https://doi.org/10.54499/UIDB/04516/2020) and ref. UIDP/04516/2020 (https://doi.org/10.54499/UIDP/04516/2020) with the financial support of FCT.IP; and Fundação para a Ciência e Tecnologia ref. 2023.03647.BDANA}
\renewcommand{\thefootnote}{\arabic{footnote}}

The large volume of surveillance video data streamed daily becomes increasingly more difficult to monitor manually. Consequently, automatic Anomaly Detection (AD) methods have become an important tool for assisting humans. Traditional deep learning AD methods have achieved strong performance through years of research. However, these methods struggle to adapt to the open-ended nature of real-world anomalies, requiring supervised fine-tuning and expensive data collection.

With the emergence of Large Vision-Language Models (LVLMs)~\cite{ref_qwen25vl}, zero-shot video understanding solutions offer strong generalization and instruction-following capabilities. However, when prompted directly for AD, these models exhibit a strong normality bias, often failing to recognize crime and softening the unusual reality depicted in anomalous surveillance videos. This limitation is exacerbated by the disparity between anomalous surveillance data and the safer, more common scenarios in LVLM training corpora. While strategies such as VERA~\cite{ref_vera} attempt to remove this bias by optimizing a fixed set of anomaly-specific questions, they still lack the flexibility to cover the long tail of real-world anomalies. This challenge also extends to advanced reasoning frameworks. 
Foundational language reasoning methods such as Chain-of-Thought (CoT)~\cite{ref_cot} encourage step-by-step deduction, and more complex approaches such as Tree-of-Thoughts (ToT)~\cite{ref_tot} explore multiple reasoning paths to improve robustness. Iteration-of-Thought~\cite{ref_iot} (IoT) addresses single-step failure by incrementally building context at each iteration. Layered Chain-of-Thought~\cite{ref_lcot} (LCoT) decomposes reasoning into independent layers that explore complementary paths and expose hallucinations through cross-layer inconsistencies. Despite their success in general problem-solving, they ultimately inherit the underlying model's normality bias when applied to AD. Consequently, their reasoning paths naturally deviate toward non-criminal aspects, failing to incorporate the inductive biases necessary for the surveillance domain.

In this work, we propose Chain-of-Anomaly Thought, a novel approach that maintains an exploratory dialogue between thought agents, aiming to detect the anomaly present in the surveillance video. This dialogue is conducted between a \textit{Witness} LVLM agent, providing a language interface for understanding the video content; a \textit{Detective} Large Language Model (LLM), responsible for generating follow-up questions; and a \textit{Supervisor} LLM that guides the reasoning process by informing the \textit{Detective} agent on the current exploration topic.

\begin{figure}[t]
    \centering
    \includegraphics[width=0.9\linewidth]{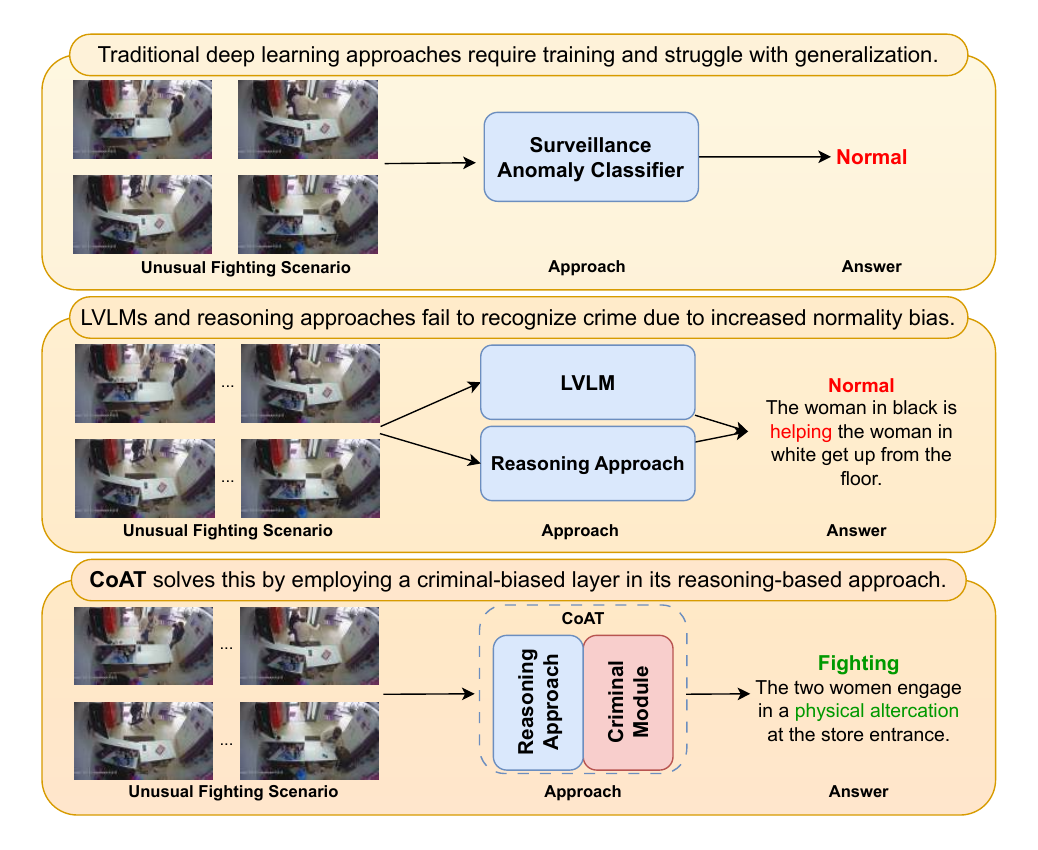}
    \caption{\textbf{CoAT’s advantage over other approaches}. Traditional methods struggle with generalization, while LVLMs and reasoning-based approaches fail due to increased normality bias. CoAT addresses these issues by employing a criminal-biased layer within its reasoning.}
    \label{fig:example}
\end{figure}


\section{Methodology}

\paragraph{Chain-of-Anomaly Thoughts (CoAT)} defines an exploratory dialogue between thought agents aiming to classify a video anomaly. The dialogue is structured in a reasoning graph that acts as a persistent memory for the current State-of-Thoughts (SoT). SoT nodes represent question-answer pairs performed within a logical reasoning path, while edges represent the operations that dictate the course of exploration. CoAT utilizes three thought agents: the \textit{Witness}, instantiated by a LVLM responsible for providing an interface for obtaining answers to questions regarding the video; the \textit{Detective} LLM, which generates pertinent questions according to the current target being explored; and the \textit{Supervisor} LLM, responsible for managing the SoT and guiding the exploration direction by instructing the \textit{Detective} agent on the current exploration goal.

\paragraph{State-of-Thoughts} structures the reasoning process into two iterative steps. First, the agents perform a non-criminally-biased structured exploration, developing themed reasoning layers such as: scenario understanding (location, time and scene objects), entity extraction (people grouping, demographics and clothing), social context (proxemics, gestures and social roles), and event understanding (actions, spatiotemporal information, causality and abnormality cues). This stage is managed by the \textit{Supervisor} through the selection of operations. The defined operations include: proceed, which instructs the \textit{Detective} to generate three pertinent follow-up questions and select the most relevant one based on the \textit{Witness}’s answers and their contribution toward the main goal; refine, which instructs the \textit{Detective} to rewrite the previous question using the additional context obtained; split, which divides a node into multiple reasoning branches; and stop, which marks a node as fully explored. 
\paragraph{Anomaly Classification Layer.} After exploration, the \textit{Witness} is prompted with a set of predefined, optimized questions tailored to induce anomaly-specific biases. The \textit{Supervisor} then outputs a final classification based on the combined evidence from both the unbiased exploration and the targeted, criminally-biased inquiry.

\begin{figure}[t]
    \centering
    \includegraphics[width=0.9\linewidth]{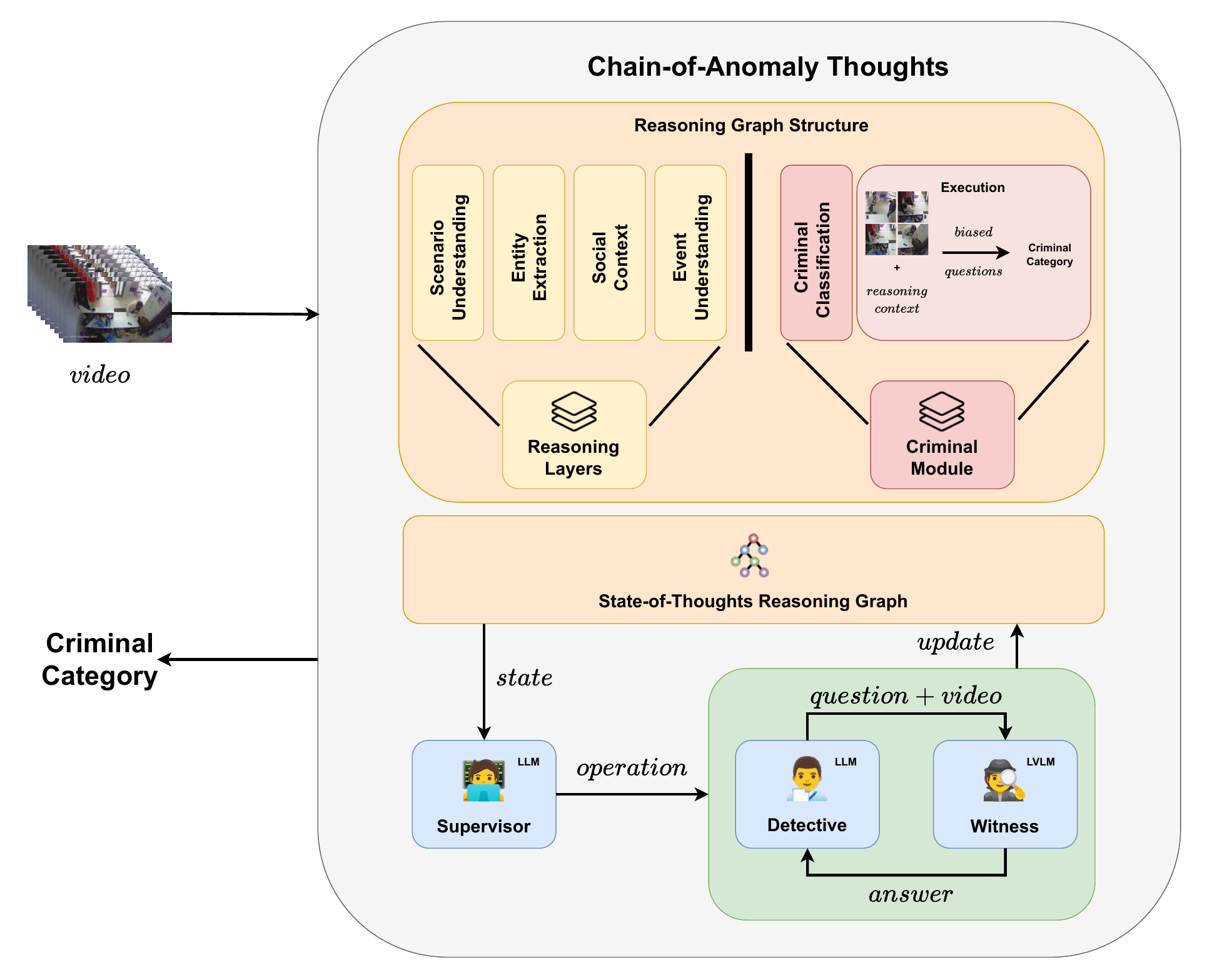}
    \caption{Overview of CoAT, a multi-agent reasoning framework for anomaly detection and classification.}
    \label{fig:example}
\end{figure}
\section{Results}

\paragraph{Experimental Setup}
In our experiments, we used Qwen2.5~\cite{ref_qwen25} (7B) and Qwen2.5-VL~\cite{ref_qwen25vl} (7B) as the baseline LLM and LVLM, respectively. Qwen2.5 offers extended context length and stronger reasoning capabilities, allowing greater sustained performance during iterative dialogue cycles, and Qwen2.5-VL is a strong multimodal baseline that incorporates time encoding, allowing it to process longer videos with event localization at second level. To evaluate the AD and Anomaly Classification (AC) tasks, we utilized the popular UCF-Crime~\cite{ref_ucfcrime} dataset, which contains 13 criminal classes. Due to the extremely low quality of some videos in UCF-Crime, we complement our testing with \textbf{BetterUCF}, a prototype dataset of high-quality videos composed of a curated selection of UCF-Crime videos with higher visual quality along with a set of videos extracted from the website ItemFix~\cite{ref_itemfix}. We ablate 4 variants of our solution, that use only a single reasoning layer. L1 refers to scenario understanding, L2 to entity extraction, L3 to social context understanding and L4 to event understanding. We also ablate a joint version combining all layers. These are always followed by the criminal layer for classification. We evaluate the task of AD, with the goal of classifying the video into Normal or Abnormal; and AC, which classifies the video into one of the categories from UCF-Crime.

\begin{figure}[t]
	\centering
	\begin{minipage}{0.24\textwidth}
		\centering
		\includegraphics[width=\linewidth]{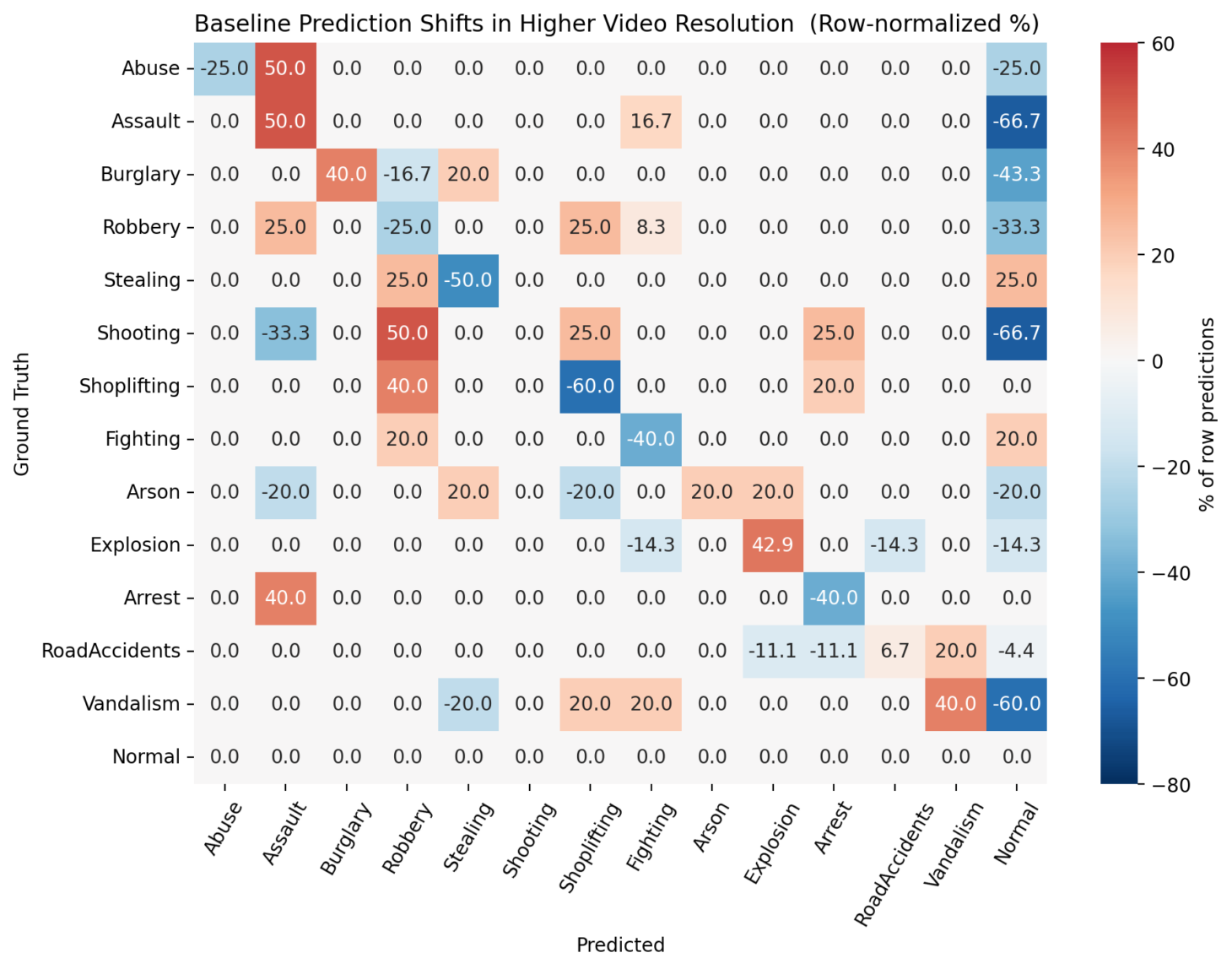}
	\end{minipage}
	\hfill
	\begin{minipage}{0.24\textwidth}
		\centering
		\includegraphics[width=\linewidth]{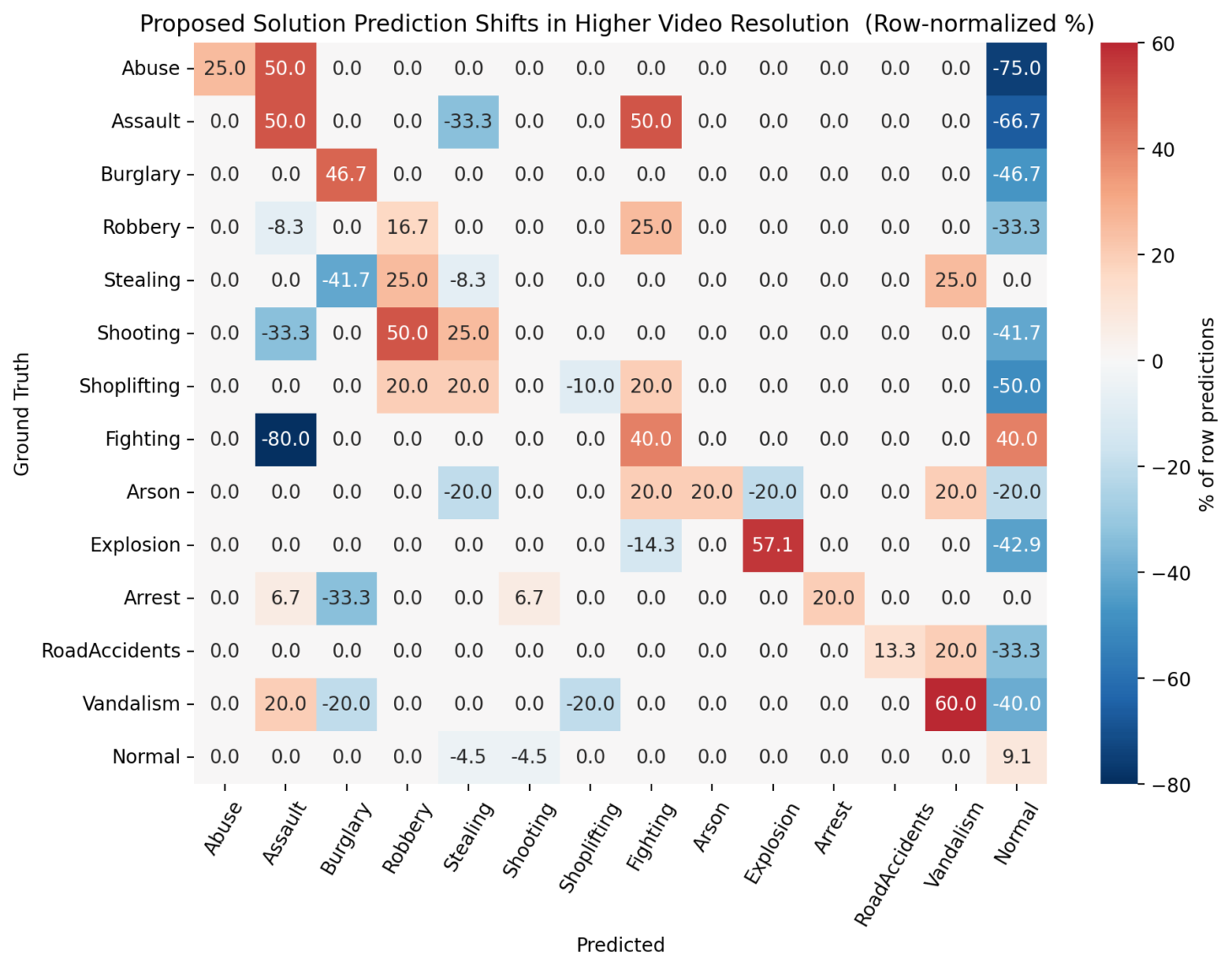}
	\end{minipage}
	\caption{\textbf{Impact of higher video resolution on the AC task}, displayed through the difference (high-resolution - low-resolution) of row-normalized confusion matrices.
		\textbf{Left}: baseline difference confusion matrix.
		\textbf{Right}: proposed solution (L4) difference confusion matrix.}
	\label{fig:resolution_impact_matrices}
\end{figure}

\paragraph{Discussion}
The F1 score comparison in Table 1 emphasizes the importance of inductive biases for AD and AC. Reasoning baselines under-perform in UCF-Crime due to the increased normality bias introduced by the exploration of non-criminal aspects, whereas our proposed solution consistently shows stronger performance through the introduction of crime-oriented biases in the criminal module. Single layer variants hold the highest results in every tested scenario with the exception of AC in the UCF-Crime dataset. Additionally, the proposed variants also achieve the majority of second and third best results, demonstrating solution robustness and effectiveness.
Our proposed solution L4 surpasses the baseline by 11.8 p.p. in the UCF-Crime AD task and by 3.78 p.p. in the BetterUCF AC task, becoming our most prominent solution variant.
We hypothesize that the joint variant suffers a performance drop due to the accumulated normality-biased context coming from all the reasoning layers. In the single-layer variants, this bias is more limited, allowing the criminal-biased questions to have a stronger corrective impact. It would also be counterintuitive to expect that combining the reasoning explorations would lead to a worse solution, further suggesting that the degradation stems from the propagation of normality bias rather than a lack of reasoning capabilities. It is also clear that resolution has a large impact on the AD and AC capabilities, as the lowest increase in AD performance is still 35.32 p.p. in solution L1. 
Figure \ref{fig:resolution_impact_matrices} displays the difference between the normalized confusion matrices of high and low resolution videos. The results show a positive difference in the diagonal of the matrix, indicating that higher resolution videos tend to be more accurately classified in comparison to low resolution or noisy videos from the same category.

\section{Conclusion}

In this work, we addressed the normality bias of LVLMs in surveillance, worsened by standard reasoning techniques. We proposed CoAT, a multi-agent reasoning framework with an anomaly-biased classification layer, achieving a 11.8 p.p. AD F1-score improvement on low-resolution data. The results highlight a key insight: exploratory reasoning can amplify normality bias, making it harmful for specialized tasks. CoAT shows the necessity of domain-specific inductive biases to achieve robust AD performance.

\begin{table}[t]
\centering
\begin{tabular}{l|cc|cc}
\hline
\textbf{Method} & \multicolumn{2}{c|}{\textbf{UCF-Crime}} & \multicolumn{2}{c}{\textbf{BetterUCF}} \\
 & A.D. & A.C. & A.D. & A.C. \\
\hline
Baseline   & 40.28 & \textbf{48.33} & 78.88 & \underline{48.11} \\
\hline
CoT        & 41.63 & 34.39 & - & -\\
ToT        & 50.56 & 35.60 & - & - \\
IoT        & 43.97 & 27.44 & - & - \\
LCoT       & \underline{51.73} & 29.34 & - & - \\
\hline
Ours (L1)  & 50.93 & 34.47 & 86.25 & \dashuline{45.76} \\
Ours (L2)  & \dashuline{51.31} & 36.45 & \textbf{88.84} & 44.34 \\
Ours (L3)  & 45.01 & 31.95 & \underline{87.99} & 41.79 \\
Ours (L4)  & \textbf{52.08} & \underline{42.68} & \dashuline{87.77} & \textbf{51.89} \\
Ours (Joint) & 50.71 & \dashuline{39.28} & 87.60 & 39.10 \\
\hline
\end{tabular}
\caption{\textbf{F1-Score (\%) comparison between baselines and CoAT.}}
\label{tab:results}
\end{table}

\bibliography{egbib}
\end{document}